# Facial Emotion Recognition:
# State of the Art Performance on FER2013

Yousif Khaireddin[1*] and Zhuofa Chen[1*]

{ykh, zfchen}@bu.edu

[1]Dept. of Electrical and Computer Engineering, Boston University, Boston, MA, USA
[*]Contributed equally to this work

**Abstract**

*Facial emotion recognition (FER) is significant for human-computer interaction such as clinical practice and behavioral description. Accurate and robust FER by computer models remains challenging due to the heterogeneity of human faces and variations in images such as different facial pose and lighting. Among all techniques for FER, deep learning models, especially Convolutional Neural Networks (CNNs) have shown great potential due to their powerful automatic feature extraction and computational efficiency. In this work, we achieve the highest single-network classification accuracy on the FER2013 dataset. We adopt the VGGNet architecture, rigorously fine-tune its hyperparameters, and experiment with various optimization methods. To our best knowledge, our model achieves state-of-the-art single-network accuracy of 73.28 % on FER2013 without using extra training data.*

**Introduction**

Facial emotion recognition refers to identifying expressions that convey basic emotions such as fear, happiness, and disgust, etc. It plays an important role in human-computer interactions and can be applied to digital advertisement, online gaming, customer feedback assessment, and healthcare [1]–[3]. With advancements in computer vision, high emotion recognition accuracy has been achieved in images captured under controlled conditions and consistent environments, rendering this a solved problem [4]. Challenges persist in emotion recognition under naturalistic conditions due to high intra-class variation and low inter-class variation, e.g. changes in facial pose and subtle differences between expressions.

Developments in computer vision continuously aim to improve classification accuracy on such problems [5]–[7]. In image classification, Convolutional Neural Networks (CNNs) have shown great potential due to their computational efficiency and feature extraction capability[8]. They are the most widely used deep models for FER [5]–[7], [9]–[11].

One specific emotion recognition dataset that encompasses the difficult naturalistic conditions and challenges is FER2013. It was introduced at the International Conference on Machine Learning (ICML) in 2013 and became a benchmark in comparing model performance in emotion recognition. Human performance on this dataset is estimated to be 65.5 % [12]. In comparing different methods and benchmarking our results, we are strictly concerned with previous work trained and evaluated on this dataset.

In this work, we aim to improve prediction accuracy on FER2013 using CNNs. We adopt the VGG network and construct various experiments to explore different optimization algorithms and learning rate schedulers. We thoroughly tune the model and training hyperparameters to achieve state-of-the-art results at a testing accuracy of 73.28 %. To our best knowledge, this is the highest single-network accuracy achieved on FER2013 without using any extra training data. We then construct several saliency maps to better understand the network's performance and decision-making process.

**Related work**

Since being introduced in the late 1990s, CNNs have shown great potential in image processing [13]. A typical CNN includes a convolutional layer, a pooling layer, and a fully connected layer. This makes it efficient in static image manipulation. However, at that time, the application of CNNs was limited due to the lack of training data and computing power. After the 2010s, the growth of computing power and the collection of larger datasets made CNNs a much more viable tool in feature extraction and image classification [8].

Various techniques have been proposed to even further improve performance. For instance, the Sigmoid activation function has been replaced by Rectified Linear Unit (ReLU) activation to avoid gradient dispersion problems and speed up training [14]. Different pooling methods such as average pooling and max pooling are used to downsample the inputs and aid in generalization [15], [16]. Dropout, regularization, and data augmentation are used to prevent overfitting. Batch normalization has been developed to help prevent gradient vanishing and exploding [17], [18].

A great deal of research has also been done in creating different optimization algorithms used in training. Though there is no systematic theoretical guideline on choosing an optimizer, empirical results show that a suitable optimization algorithm can effectively improve a model's performance [19]. The most commonly used optimizer is Stochastic gradient descent (SGD). It is a simple technique that updates the parameters of a model based on the gradient of a single data point [19]. Numerous variations of this algorithm have been proposed to speed up training. AdaGrad adaptively scales the learning rate for each dimension in the network [20]. RMSProp radically diminishes the learning rate [21]. Adam combines the advantages of AdaGrad and RMSProp by scaling the learning rate and introducing momentum of gradient, etc. [22].

Among many others, one significant factor that could impact performance is the learning rate. A large learning rate could lead to oscillations around the minima or divergence in the loss. A small learning rate would slow down the model's convergence significantly and could trap the model at a non-optimal local minimum. A commonly used technique is to employ a learning rate scheduler that changes the learning rate during training [23]. For instance, time-based decay reduces the learning rate either linearly or exponentially as the iteration number increases. Step decay drops the learning rate by a factor after certain epochs. An adaptive learning rate schedule tries to automatically adjust the learning rate based on the local gradients during training. Cosine annealing resets the learning rate periodically and reuses "good weights" during the training process, etc. [24], [25].

With all the developments above and the extensive research in CNNs, they have become an extremely favorable tool when tackling tasks in image processing, pattern recognition, and feature extraction. Once a large facial expression dataset, FER2013, was introduced at ICML in 2013, it became a benchmark in comparing model performance in emotion recognition. Many CNN variants have achieved remarkable results with a classification accuracy between 65 % and 72.7 % [26]–[34]. For instance, Liu et. al. trained the three separate CNNs and ensembled them to improve performance. Their best single-network accuracy is 62.44 % [26]. Minaee et. al used an attentional convolutional network in an end-to-end deep learning framework and achieved an accuracy of 70.02 % [27]. Tang et. al. replaced the softmax layer with a support vector machine in a deep neural network and achieved a classification accuracy of 71.2 % [32]. Shi et. al. proposed a novel amend representation module (ARM) to substitute the pooling layer and achieved a testing accuracy of 71.38 % [33]. Pramerdorfer et. al. compared the performance of three different architectures, VGG, Inception, and ResNet. Their results show that VGG performs best at an accuracy of 72.7 %, followed by ResNet at 72.4 %, and Inception at 71.6 % [34].

Ensembling multiple different models has been shown to improve performance. For instance, Liu et. al. ensembled of 3 CNNs and improved their accuracy by 2.6 % [26]. Pramerdorfer et. al. ensembled 8 CNNs and achieved a 2.5 % performance boost [34]. However, in order to improve the ensemble performance even further, we aim to first optimize the building blocks of these ensembles, a single network. Other research work has tried to improve their model's performance on FER2013 by including auxiliary training data; however, that is out of the scope of this paper.

# Experiments

## A. Dataset, Preprocessing, and Augmentation

In training on FER2013, we adhere to the official training, validation, and test sets as introduced by the ICML. FER2013 consists of 35888 images of 7 different emotions: anger, neutral, disgust, fear, happiness, sadness, and surprise. A Kaggle forum discussion held by the competition organizers places human accuracy on this dataset in the range of 65 – 68 % [32].

To account for the variability in facial expression recognition, we apply a significant amount of data augmentation in training. This augmentation includes rescaling the images up to ± 20 % of its original scale, horizontally and vertically shifting the image by up to ± 20 % of its size, and rotating it up to ± 10 degrees. Each of the techniques is applied randomly and with a probability of 50 %. After this, the image is then ten-cropped to a size of 40×40, and random portions of each of the crops are erased with a probability of 50 %. Each crop is then normalized by dividing each pixel by 255.

## B. Training and Inference

We run all experiments for 300 epochs optimizing the cross-entropy loss. In the following sections, we vary the optimizer used as well as the learning rate schedulers and maintain other parameters constant. We use a fixed momentum of 0.9 and a weight decay of 0.0001. All experiments are run with gradient scaling to prevent gradient underflow. The models are evaluated using validation accuracy and tested using standard ten-crop averaging.

## C. VGGNet Architecture

VGGNet is a classical convolutional neural network architecture used in large-scale image processing and pattern recognition [35]. Our variant of VGGNet is shown in Figure 1. The network consists of 4 convolutional stages and 3 fully connected layers. Each of the convolutional stages contains two convolutional blocks and a max-pooling layer. The convolution block consists of a convolutional layer, a ReLU activation, and a batch normalization layer. Batch normalization is used here to speed up the learning process, reduce the internal covariance shift, and prevent gradient vanishing or explosion [18]. The first two fully connected layers are followed by a ReLU activation. The third fully connected layer is for classification. The convolutional stages are responsible for feature extraction, dimension reduction, and non-linearity. The fully connected layers are trained to classify the inputs as described by extracted features.

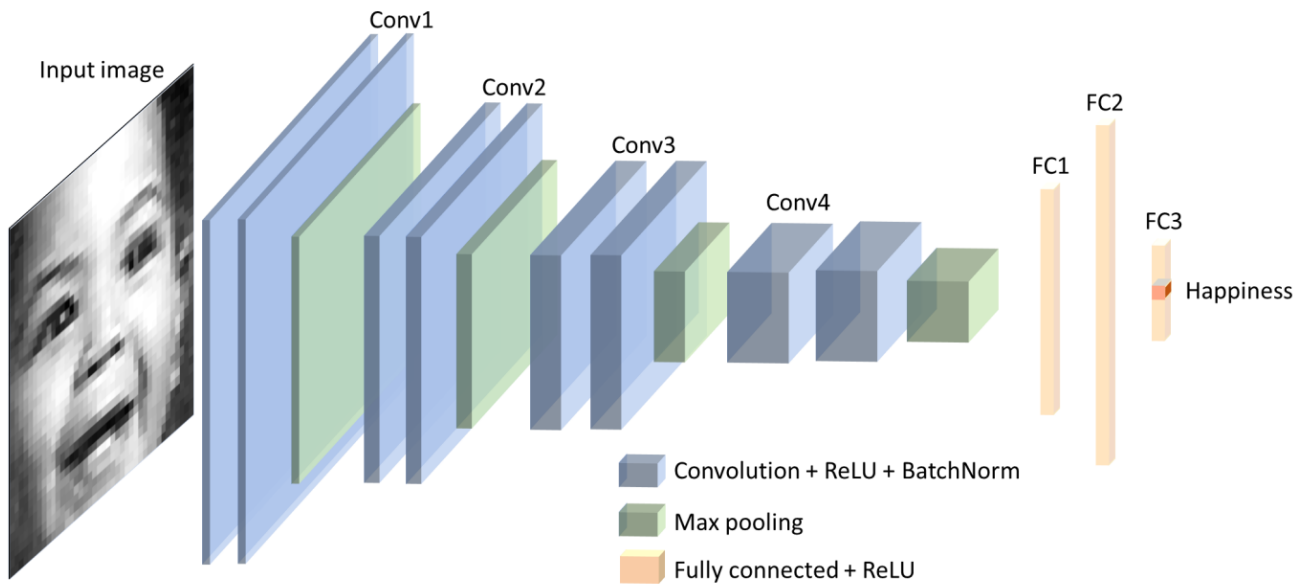

*Figure 1 VGGNet architecture. A face expression image is fed into the model. The four convolutional blocks (Conv) extract high-level features of the image and the fully-connected (FC) layers classify the emotion of the image.*

## D. Tuning

Initially, we tune our model architecture to maximize performance. All initial experiments were run using SGD. A grid search is performed to determine the optimal batch size and the best drop-out rate after our fully connected layers. Once the architecture has been optimized, we then explore the effects of different optimizers and learning rate schedulers on our model's performance. We then set up a final experiment to try and fine-tune the trained model's weights and improve its performance.

### D.1 Optimizer

Our first experiment intends to find the best optimizer in training our architecture. For this, we explore 6 different algorithms: SGD, SGD with Nesterov Momentum, Average SGD, Adam, Adam with AMSGrad, Adadelta, and Adagrad. Although many of these algorithms are very closely related, understanding how they perform differently in this optimization will help us understand the importance of their subtle differences.

We run 2 different variations of this experiment. In the first variation, we run all algorithms with a fixed learning rate of 0.001. This learning rate was determined using a grid search. In the second variation, we set up a simple learning rate scheduler with an initial learning rate of 0.01 and it is reduced by a factor of 0.75 if the validation accuracy plateaus for 5 epochs. The parameters of this scheduler were also determined using a grid search. All other parameters, such as weight decay, momentum, dropout, and batch size are maintained constant after the initial optimization.

### D.2 LR Schedule

The next experiment we run is to find the optimal learning rate scheduler. In this section, we run the same architecture, using the optimal optimizer decided by the previous section with 5 different schedulers: Reduce Learning Rate on Plateau (RLRP), Cosine Annealing (Cosine), Cosine Annealing with Warm Restarts (CosineWR), One Cycle Learning Rate (OneCycleLR), and Step Learning Rate (StepLR). For a baseline, we also run a model using a constant learning rate that was determined using a grid search. All schedulers have an initial learning rate of 0.01 and their parameters are selected using a grid search. All other parameters are maintained constant.

Although these schedulers have their similarities, they perform the learning rate update differently. For example, StepLR and RLRP both have a similar assumption that the longer we train, the smaller our step sizes should be. However, StepLR systematically reduces the learning rate after a set number of epochs while RLRP monitors the model's current performance before making a learning rate update. Similarly, both Cosine Annealing and Cosine Annealing with Warm Restarts have a cosine-based learning rate update function. So, the learning rate gradually oscillates between two values. However, the key difference is that the latter one resets the model's parameters regularly to maintain good model weights before taking update steps. These subtle differences could drastically change the resulting performance.

### D3. Fine Tuning

To further increase our model's accuracy, we then experimented with hyper-tuning the final weights of our model. We reload the parameters and train for a final 50 epochs using an initial learning rate of 0.0001. We set this learning rate to maintain the update steps small; thus, ensuring that our model's weights are not skewed far away, since this is an already trained model. This experiment is run using Cosine Annealing and Cosine Annealing with Warm Restarts since both of these schedulers slowly oscillate the learning rate back and forth thus not allowing for major weight changes. The second schedule is also favorable since its warm restarts would mean that it would regularly reset the model's weights back to some good location during its updates.

We then ran a second variation of this experiment in which we combined the validation set into training to allow for a larger dataset set when tuning. This larger dataset would allow the model to have more samples to learn from; thus, improving its performance. The test set is left un-altered and all other parameters are kept constant.

By running two variations of this experiment, we can verify two things. Using the first variation, we can confirm the effectiveness of the tuning. Using the second experiment, we can understand the benefits of added data.

## Results

### A. Optimizer

We first study and compare the effect of optimizers on the performance of our model. Figure 2 shows the validation accuracy attained by our model using the different optimizers. The yellow bars show the first variation of this experiment with a constant learning rate and the orange bars show the second variation using a decaying learning rate. Excluding Adadelta, all optimizers show a high validation accuracy, above 70 %. The model using the SGD with Nesterov momentum performs the best in both experiments attaining a validation accuracy of 73.2 % and 73.5 %. We also found that Adam and its AMSGrad variant perform better than Adadelta and Adagrad. This is because Adam optimization combines the advantage of AdaGrad and RMSProp by introducing momentum of the gradient. Finally, on this dataset, all variants of SGD are outperforming all other optimizers.

### B. LR Schedule

Next, we explore the effects of different learning rate schedulers on our model. Figure 3 shows the validation and testing accuracies attained by our models. All runs here use the best performing optimizer based on the previous section, SGD with Nesterov momentum. The first thing to note is that Reducing Learning Rate on Plateau (RLRP) performs best. It achieves a validation accuracy of 73.59 % and a testing accuracy of 73.06 %. To our best knowledge, this is already surpassing the previous single-network state-of-the-art performance.

For the next set of comparisons, we will be focused on the validation accuracy, since the testing accuracies we are reporting are strictly for benchmarking publicly. The constant learning rate outperforms some of the other schedulers (OneCycleRL and StepLR). For OneCycleLR, this could be since it is usually intended for fast training with larger learning rates and this may not be applicable on FER2013 [36]. Cosine Annealing and its counterpart Cosine Annealing with Warm Restart perform comparably. Comparing StepLR and RLRP, they both slowly reduce the learning rate to a minimum. RLRP performs better since it monitors the current performance before deciding when to drop the learning rate as opposed to systematically reducing the learning rate.

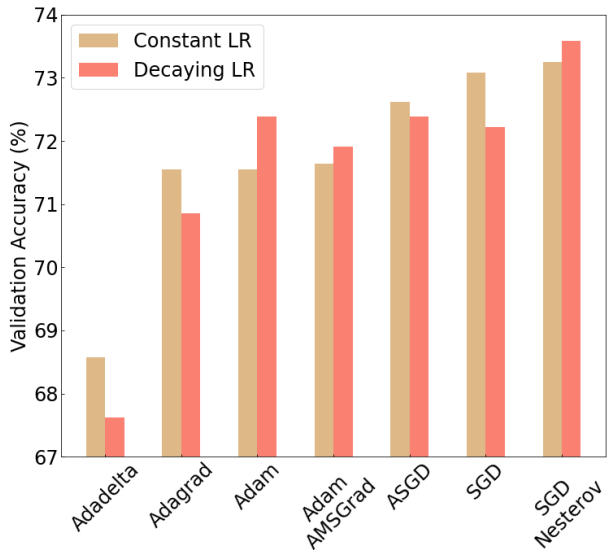

*Figure 2 VGGNet performance using different optimizers. The yellow bars show the results using a constant learning rate (LR). The orange bars show accuracies using decaying LR.*

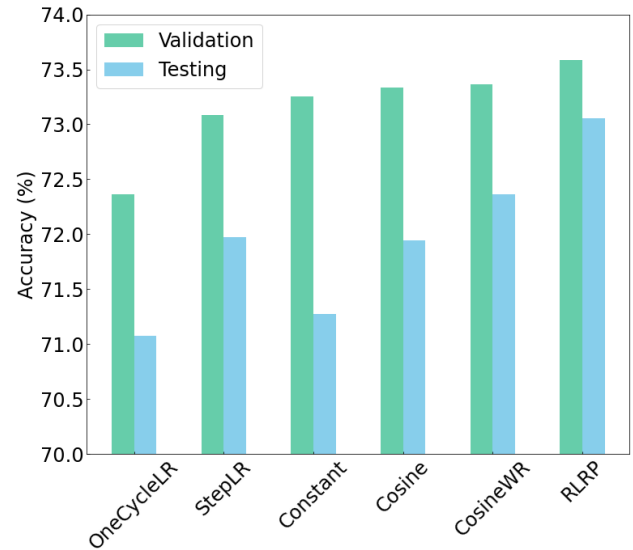

*Figure 3 VGGNet performance using different LR schedulers. The green bars show the final validation accuracies and the blue ones show the corresponding testing accuracies.*

## C. Fine Tuning

To further increase our final performance, we then run an experiment in which we reload our best performing model thus far and tune it for 50 extra epochs using both cosine annealing schedulers at a small learning rate. We then also combine our training and validation data and perform this same tuning in a separate run. Table 1 shows the final testing accuracies achieved in this experiment benchmarked against the model which they are tuning. Cosine Annealing performs best here and improves the model by 0.05 %. Cosine Annealing with Warm Restarts, on the other hand negatively impacts performance in this tuning stage reducing the accuracy by 0.42 %. As expected both models perform better after training on the combined dataset resulting from the training and validation data. Our final best model achieves an accuracy of 73.28 %. Once again, the testing data is left unaltered at all stages.

*Table 1 Testing accuracy after extra tuning.*

| Methods | | Testing Accuracy |
|---|---|---|
| Trained VGGNet | | 73.06 % |
| Regular split | Cosine + WR | 72.64 % |
| | Cosine | 73.11 % |
| Combine training and validation | Cosine + WR | 73.14 % |
| | **Cosine** | **73.28 %** |

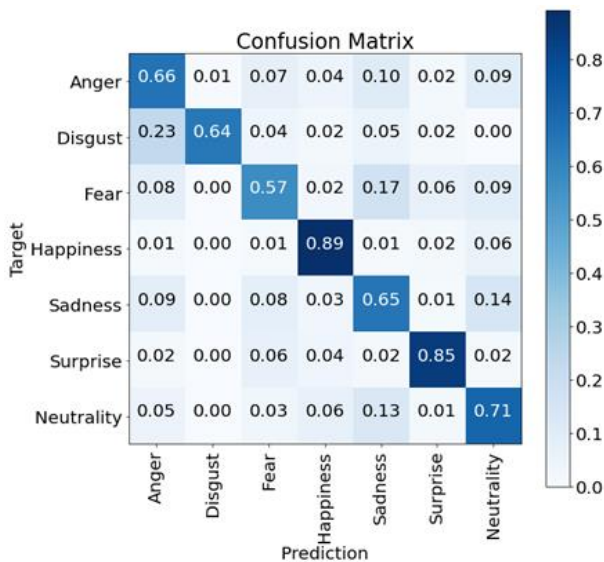

*Figure 4 The confusion matrix of our final VGGNet on the FER2013 public test set.*

## D. Confusion Matrix

Figure 4 shows the final model's confusion matrix on the FER2013 testing set. The model shows the best classification on the "happiness" and "surprise" emotions. On the other hand, it makes the most mistakes when classifying between "disgust" and "anger". Next, the low classification accuracy in "disgust" and "fear" can be attributed to the fact that they have a lower number of samples in the original training set. The misclassification between "fear" and "sadness" may be due to the inter-class similarities of the dataset

## E. Performance Comparison

Table 2 summarizes previous reported classification accuracies FER2013. Most reported methods perform better than the estimated human performance (~ 65.5 %). The previous best-reported single-network accuracy is 72.7 % [34]. In this work, we achieve the state-of-the-art accuracy of 73.28 %.

*Table 2 Fer2013 public testset benchmark.*

| Method | Accuracy Rate |
|---|---|
| CNN [26] | 62.44 % |
| GoogleNet [30] | 65.20 % |
| VGG+SVM [29] | 66.31 % |
| Conv + Inception layer [31] | 66.40 % |
| Bag of Words [28] | 67.40 % |
| Attentional ConvNet [27] | 70.02 % |
| CNN + SVM [32] | 71.20 % |
| ARM (ResNet-18) [33] | 71.38 % |
| Inception [34] | 71.60 % |
| ResNet [34] | 72.40 % |
| VGG [34] | 72.70 % |
| **VGG (this work)** | **73.28 %** |

## F. Saliency map

The 'black box' nature of deep learning models makes it difficult to understand how the model processes the input image to obtain the final result. Visualizing the information captured inside deep neural networks is important in evaluating the effectiveness of the model and understanding how it computes its prediction. Thus, generating an understandable visualization of our CNN can help describe how it differentiates between and captures different facial emotions.

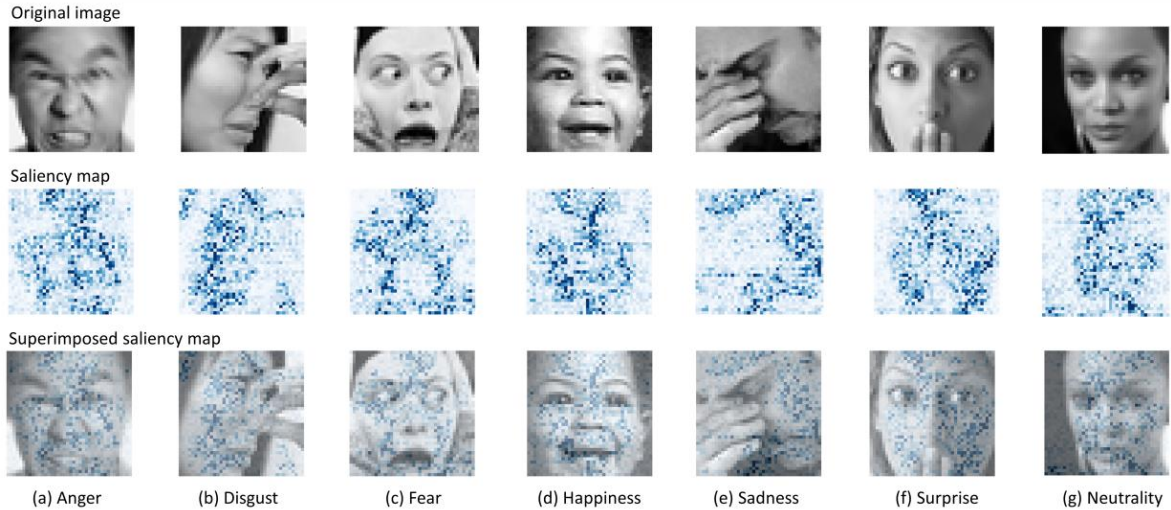

*Figure 5 Original images, Saliency maps, and superimposition for all emotions in FER2013.*

One of the common visualization techniques in deep neural networks is called a saliency map [37]. By propagating the loss back to the pixel values, a saliency map can highlight the pixels which have the most impact on the loss value. It highlights the visual features the CNN can capture from the input; thus, allowing us to better understand the importance of each feature in the original image on the final classification decision.

We generate the saliency map using our best performing network to understand how it classifies each emotion in the FER2013 dataset. Figure 5 shows a saliency map for each emotion before and after being superimposed on the image it is generated from. Judging by all the images, our CNN can effectively capture most of the critical regions for each emotion. The model is placing a large importance on almost all facial features of the person in each image. This is most clearly seen in (f) where the saliency map almost perfectly maps the entire face of the woman. Our model is also effectively dropping regions like the background in (a), (d) and (g), the hair in (a), (c), and (g), and the hand in (e), all of which are not very informative when it comes to describing someone's emotion.

There are some clear mistakes in the saliency maps, this can be seen in (b), (e), and the corner of (g) where the model highlights some of the background pixels. We think that a model that can more effectively identify the facial features in an image and drop all useless information will perform better on this dataset.

**Conclusion**

This paper achieves single-network state-of-the-art classification accuracy on FER2013 using a VGGNet. We thoroughly tune all hyperparameters towards an optimized model for facial emotion recognition. Different optimizers and learning rate schedulers are explored and the best initial testing classification accuracy achieved is 73.06 %, surpassing all single-network accuracies previously reported. We also carry out extra tuning on our model using Cosine Annealing and combine the training and validation datasets to further improve the classification accuracy to 73.28 %. For future work, we plan to explore different image processing techniques on FER2013 and investigate ensembles of different deep learning architectures to further improve our performance in facial emotion recognition.